\documentclass[letterpaper, 10 pt, conference]{IEEEtran}

\IEEEoverridecommandlockouts                              % This command is only needed if 
\usepackage{mathptmx} % assumes new font selection scheme installed
\usepackage{bm}
\usepackage{times} % assumes new font selection scheme installed
\usepackage{amsmath} % assumes amsmath package installed
\DeclareMathOperator*{\argmax}{arg\,max}
\DeclareMathOperator*{\argmin}{arg\,min}
\usepackage{amssymb}  % assumes amsmath package installed
\usepackage{graphicx}
\usepackage{mathtools}
\usepackage{upgreek}
\usepackage{gensymb}
\usepackage{booktabs}

\ifCLASSOPTIONcompsoc
    \usepackage[caption=false, font=normalsize, labelfont=sf, textfont=sf]{subfig}
\else
\usepackage[caption=false, font=footnotesize]{subfig}
\fi
\usepackage[noadjust]{cite}
\usepackage{placeins}
\usepackage{xcolor}
\usepackage{url}
\usepackage{mwe}
\usepackage{balance}
\usepackage{multirow}

\DeclarePairedDelimiter\abs{\lvert}{\rvert}%
\DeclarePairedDelimiter\norm{\lVert}{\rVert}%

\makeatletter
\let\oldabs\abs
\def\abs{\@ifstar{\oldabs}{\oldabs*}}
\let\oldnorm\norm
\def\norm{\@ifstar{\oldnorm}{\oldnorm*}}
\makeatother

\title{\LARGE \bf
Optimal Motion Scaling for Delayed Telesurgery
}

\author{Jason Lim$^1$, Florian Richter$^1$, Zih-Yun Chiu$^1$, Jaeyon Lee$^2$, Ethan Quist$^2$, Nathan Fisher$^2$, Jonathan Chambers$^2$, \\ Steven Hong$^3$, Michael C. Yip$^1$ \IEEEmembership{Senior Member, IEEE}
\thanks{$^1$ J. Lim, F. Richter, Z.-Y. Chiu, and M.C. Yip are with the Electrical and Computer Engineering Department, University of California San Diego, La Jolla, CA 92093 USA. {\tt\footnotesize\{jkl009, frichter, zchiu, yip\}@ucsd.edu}
\newline
$^2$ J. Lee, E. Quist, N. Fisher, and J. Chambers are with Telemedicine and Advanced Technology Research Center (TATRC), Fort Detrick, MD 21702 USA. {\tt\footnotesize\{jaeyeon.lee5.ctr, ethan.t.quist.civ, nathan.t.fisher3.civ, jonathan.m.chambers14.ctr\}@health.mil}
\newline
$^3$ S. Hong is with Atrium Health Wake Forest, Winston-Salem, NC 27157 USA. {\tt\footnotesize\{steven.Hong@atriumhealth.org\}}
}%
}

\begin{document}

\maketitle
\thispagestyle{empty}
\pagestyle{empty}

%%%%%%%%%%%%%%%%%%%%%%%%%%%%%%%%%%%%%%%%%%%%%%%%%%%%%%%%%%%%%%%%%%%%%%%%%%%%%%%%

\begin{abstract}
Robotic teleoperation over long communication distances poses challenges due to delays in commands and feedback from network latency. One simple yet effective strategy to reduce errors and increase performance under delay is to downscale the relative motion between the operating surgeon and the robot. The question remains as to what is the optimal scaling factor, and how this value changes depending on the level of latency as well as operator tendencies. We present user studies investigating the relationship between latency, scaling factor, and performance. The results of our studies demonstrate a statistically significant difference in performance between users and across scaling factors for certain levels of delay.
These findings indicate that the optimal scaling factor for a given level of delay is specific to each user, motivating the need for personalized models for optimal performance. We present techniques to model the user-specific mapping of latency level to scaling factor for optimal performance, leading to an efficient and effective solution to optimizing performance of robotic teleoperation and specifically telesurgery under large communication delay.
\end{abstract}

\section{Introduction}

There are many challenges that lead to inadequate medical care and worse surgical outcomes in rural and remote communities, as well as conflict and military scenarios.
Limited resources and a lack of highly trained care providers on site mean that patients have to travel long distances to receive adequate care, greatly increasing the chances of complications \cite{morgan2020emergency}.
One study found that the risk of death related to trauma more than doubled in the rural population, with the risk increasing with increased remoteness \cite{fatovich2011comparison, fatovich2011major}.
Similarly, limited resources and dangerous conditions provide huge challenges in delivering quality surgical care in military conflict zones \cite{wren2020consensus, ferreira2024surgical, giannou2014war, hall2020current, anagnostou2020practicing, de2022robotic}. 
Bringing doctors into remote or conflict areas is not practical as it incurs significant costs and risks to the doctors \cite{cazac2014telesurgery}.

Robotic telesurgery is a very promising technology to bridge this gap.
In 2001 Marescaux et al. demonstrated the feasibility of telesurgery by successfully performing a procedure remotely from New York to Strasbourg (about 14,000 km) over a dedicated ATM (Asynchronous Transfer Mode) network, with a mean round-trip delay of 155 ms \cite{marescaux2001transatlantic}. Although this procedure was a landmark success for long-distance telesurgery, ATM technology is expensive and cannot be practically implemented on a widespread scale \cite{cazac2014telesurgery}.
In 2003 a telerobotic surgical system was set up between St. Joseph's Hospital in Hamilton, Ontario and North Bay General Hospital 400 km away using a commercially available IP/VPN network with multiprotocol label switching technology. Over the next few years, dozens of successful procedures were completed with a typical round-trip delay ranging between 135-150 ms \cite{anvari2005establishment}. In another study, 108 animal procedures were peformed teleoperatively over a distance of 3000 km, with delay ranging from 170 to 300 ms \cite{wang2024influence}.

\begin{figure}
    \centering
    \includegraphics[width=\linewidth]{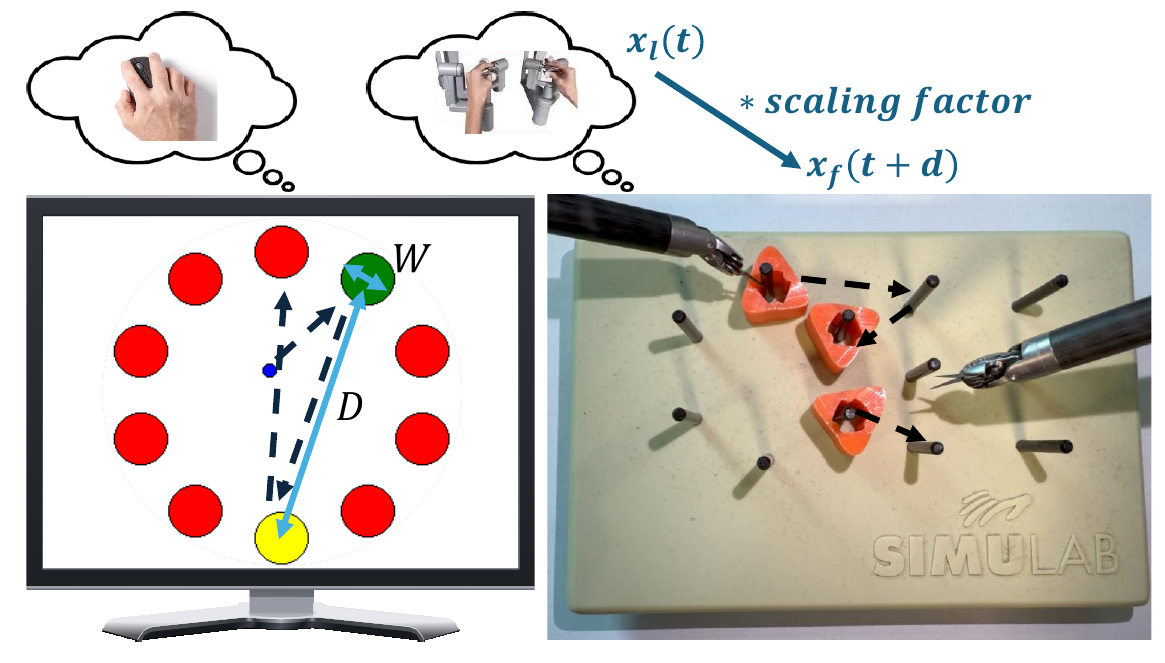}
    \caption{Motion scaling scales the leaders motion, $x_l(t)$, to improve precision when operating under delay.
    We conduct user studies in a simulated environment with a computer mouse and monitor (left), and a physical environment with the da Vinci Research Kit telesurgical system (right). Data collected from these user studies reveal that the effect of scaling factor on performance varies with latency level and among users. This data is utilized to build user-specific models that predict the optimal scaling factor for a given latency.}
    \label{fig:cover_fig}
\end{figure}

While several other researchers have reported successful telesurgery cases, the problems caused by communication delays still pose a major challenge to making telesurgery practical and widespread. Extensive studies have shown that teleoperative task completion time and error rates increase with increasing latency \cite{anvari2005impact, xu2014determination}. Reports vary in what latency levels are considered unsafe. Anvari discovered that surgeons are able to adapt to delays up to 150 ms with very little conscious effort, but found it significantly more challenging and had to consciously slow down at levels greater than 175-200 ms \cite{anvari2007remote}. In Wang's study, surgeons reported significantly increased difficulty when latency rose above 270 ms, and the authors identified 320 ms as a practical upper limit for acceptable telesurgery \cite{wang2024influence}. Some studies report impairments beginning with as little as 50 ms of delay \cite{barba2022remote}.

Many sophisticated methods have been proposed for handling latency during robotic teleoperation; one extremely straightforward solution is \textit{motion scaling}, in which a scaling factor is applied to the relative motion between leader and follower.
Suppose the leader pose at time is given by $\mathbf{x}_l(t)$, then the follower's pose $\mathbf{x}_f$ at time $t+d$ is given by
\begin{equation}    
     \mathbf{x}_f(t + d) = s * (\mathbf{x}_l(t) - \mathbf{x}_l(t-1)) + \mathbf{x}_f(t+d-1),
\end{equation}
where $s$ is the scaling factor and $d$ is the delay.
Multiple studies have clearly demonstrated that decreasing the scaling factor $s$ leads to increased accuracy and reduced error in teleoperative tasks \cite{jacobs2003limitations, prasad2004surgical, cassilly2004optimizing, richter2019motion, richter2021bench, orosco2021compensatory}.
However, it is still not known what the optimal scaling factor should be, whether it depends on the specific user, or how it should change under variable delay.

% \begin{figure*}
%     \includegraphics[width=\textwidth]{Figures/test_cover_figure3.png}
%     \caption{Motion scaling implemented on a teleoperation system scales the leader motion $x_l(t)$ to produce the follower's motion $x_f(t+d)$. We conduct user studies in a simulated environment with a computer mouse and monitor (left), as well as physical environment with the da Vinci Research Kit telesurgical system (right). Data collected these user studies reveal that the effect of scaling factor on performance varies with latency level and among users. This data can be utilized to build user-specific models that predict the optimal scaling factor for a given latency.}
%     \label{fig:cover_fig}
% \end{figure*}

\subsection{Contributions}

As a step towards making telesurgery more feasible, we investigate and model the effect of scaling factor on teleoperative performance as it varies with both level of delay and across different users. The intuition is that an optimal scaling factor will exist that varies with delay and user tendencies.
Through user study experiments in simulated and physical environments, as depicted in Fig. \ref{fig:cover_fig}, we confirm this hypothesis and show that the optimal scaling factor with respect to several performance metrics will vary with level of delay, and while general trends can be observed, different users will prefer different scaling factors per delay. 
Using the data collected from these experiments, we build personalized models that can be used to predict optimal scaling factors for specific users at any level of delay.
These results will lead to motion scaling solutions that can be easily implemented with current telesurgical technology at very little cost, and have the potential to vastly improve the quality and access to healthcare on a global level.

\subsection{Related Works}
% Going to have 4 main paragraphs: supervisory control, predictive methods, force-feedback control, and then motion scaling

% prioritize the motion scaling stuff

% Combine supervisory control and predictive display paragraphs
% shorten force-feedback paragraph
% keep or even expand upon motion scaling

% Early research into teleoperation under delay, supervisory control and  Predictive Aiding methods (predictive displays
While the body of research regarding solutions for robotic teleoperation under delay is vast and cannot be fully covered in this paper, we will address the main ideas.
One of the earliest approaches was supervisory control, in which the human operator inputs high level commands which are translated to the remote system, where sensors can be used to close the feedback loop locally with no delay and execute the given commands autonomously \cite{ferrell1967supervisory, sheridan1992telerobotics, hodvzic2015teleoperation, blackmon1996model}.
Predictive aids are another widely explored solution for teleoperation under significant delays, in which  real-time visual or haptic predictions of the robot's future state are generated based on current commands and environmental models \cite{hirzinger1989predictive, bejczy1990predictive, sheridan1993space, richter2019augmented}.
While both supervisory control and predictive display methods showed promise in reducing errors and improving performance, their reliance on accurate environment models render them unsuitable for unstructured surgical settings, where complex, unpredictable tool-tissue interactions and the demand for precise, real-time decision-making exceed the capabilities of pre-programmed autonomy.

% Force-feedback methods
Perhaps the largest body of work regarding teleoperation under delay focuses on bilateral control with force feedback. Bilateral control can suffer from instability under delay, and a plethora of techniques have been studied to mitigate this instability \cite{arcara2002control, varkonyi2014survey}; however, currently deployed telesurgical systems do not offer the force feedback necessary to implement these techniques. Additionaly, certain research suggests employing haptic feedback leads to increased completion time \cite{yip2011performance}, and we therefore seek solutions that do not involve force feedback.

% Previous research on motion scaling

One straightforward way to address delay during teleoperation is \textit{motion scaling}, which applies a scaling factor to the relative motion between leader and follower. Reducing this factor forces slower operator movements, which improves precision and accuracy, and reflects the natural 'move and wait' strategy observed by Ferrell \cite{ferrell1965remote}. The effectiveness of motion scaling was first demonstrated in non-delayed telesurgical settings, leading to decreased errors and increased accuracy \cite{jacobs2003limitations, prasad2004surgical, cassilly2004optimizing}.

Richter and Orosco built on these findings by demonstrating that motion scaling effectively reduced errors without substantially increasing completion time \cite{richter2019motion}. Further research by Orosco et al. \cite{orosco2021compensatory} investigated the effects of different scaling factors under varying levels of delay, finding that performance was relatively unaffected by scaling factor at no delay, but both error and task completion time were both improved at lower scaling factors under a delay of 750 ms. Richter et al. \cite{richter2021bench} confirmed similar benefits in live pig telesurgery using the dVRK platform. Together, these studies highlight motion scaling as a powerful means to enhance telesurgical performance under delay. 
In this paper, we expand upon this idea by proposing a method for optimizing the scaling factor at various levels of delay for specific users in order to achieve the best performance for each individual under realistic variable delay conditions.

\section{Methods}

\begin{figure} 
    \centering
  \subfloat[User A\label{user_a}]{%
       \includegraphics[width=0.99\linewidth]{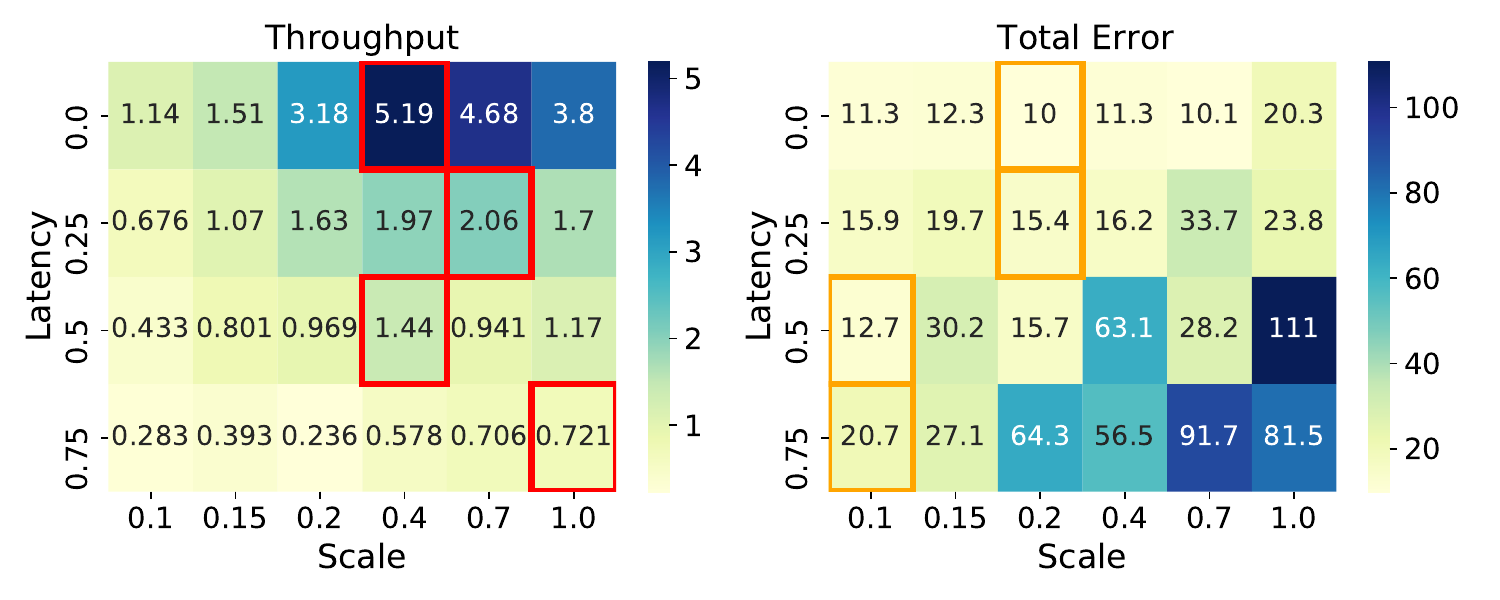}}
    \\
  \subfloat[User B\label{user_b}]{%
        \includegraphics[width=0.99\linewidth]{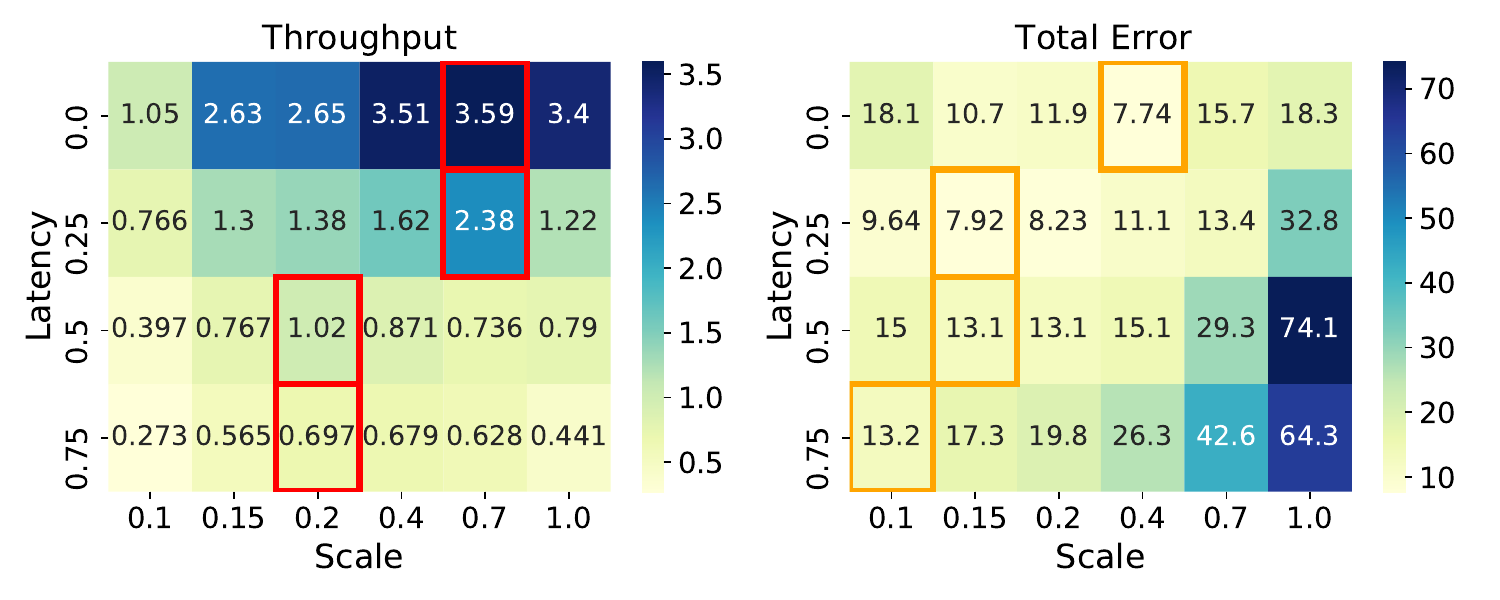}}
  \caption{Heatmaps of the results from the 2D user study illustrate the two key performance metrics of throughput and total error for two representative users, with red boxes outlining the optimal motion scaling factor for each latency level. The optimal values vary with latency and also between users, underscoring the individual-specific nature of optimal motion scaling in telesurgical tasks.}
  \label{fig:key_metrics_sim} 
\end{figure}

In order to model the relationship between performance, scaling factor, and delay, a set of experiments is conducted to gather data on how well different users perform a teleoperative task under various levels of delay and scaling factors.
Let $(P, s, d, o)$ be the collected data, where $P$ represents a set of performance metrics, $s$ the scaling factor, $d$ the level of delay, and $o$ designates the specific user. Fig. \ref{fig:key_metrics_sim} depicts an example of datasets for two metrics for two different users.
From this data, we derive personalized operator models, $\hat{P}_o(s,d)$, which predict performance from a scaling factor and delay.
We utilize personalized operator models since prior work found that latency impacts operators in different ways \cite{anvari2005impact}.
% Using the data from these experiments we can generate personalized models for each user that predicts performance for any scaling factor and level of delay in their respective domains.
These personalized models are then used to solve for the optimal scaling factor depending on the delay.

\begin{figure}[t]
    \centering
    \includegraphics[width=\linewidth]{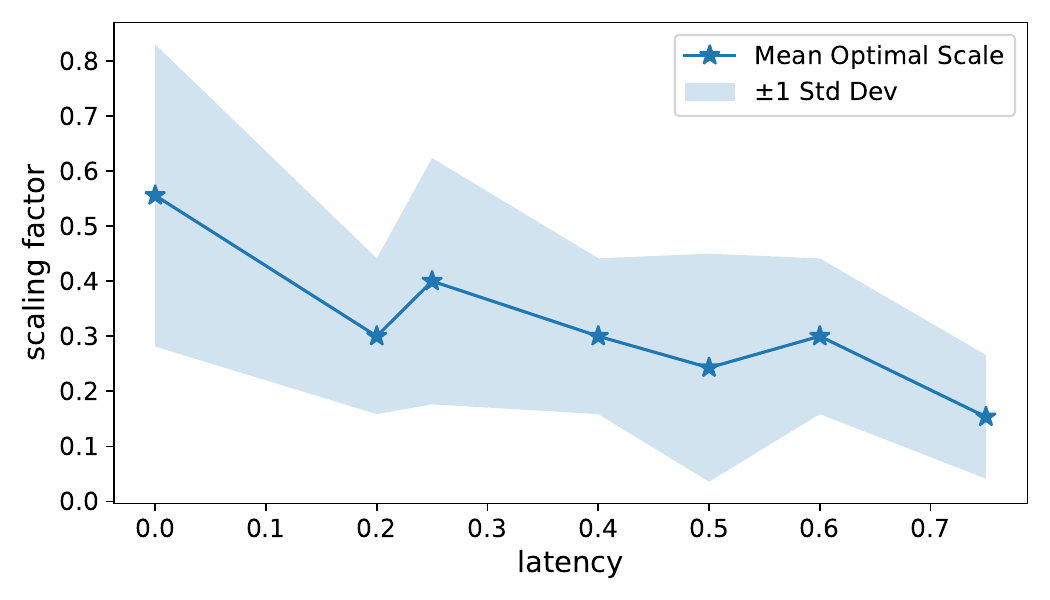}
    \caption{A plot of the mean and one standard deviation shown for optimal scaling factor as measured by the weighted performance metric demonstrates that while the optimal scaling factor varies among users for a certain latency, the general trend is for it to decrease as latency increases.}
    \label{fig:optimal_scale_vs_latency}
\end{figure}

\subsection{Operator Performance Modeling}
% Datasets consisting of variables $(P, s, d, o)$ will be collected from a series of user study experiments, where $P$ represents a set of performance metrics, $s$ the scaling factor, $d$ the level of delay, and $o$ designates the specific user.
A Bayesian Polynomial Regression (BPR) model will be used to estimate the performance metrics, $\hat{P}_o(s, d)$, as a function of scaling factor and delay.
The choice of this model was made after observing that the data seemed to follow a quadratic trend in $s$ for a given $d$, and therefore could be represented as a polynomial surface.
Also, we sought a model that captures uncertainty in predictions for developing interactive algorithms to calibrate models.
In Bayesian Regression, the outputs are assumed to be governed by a linear function with additive noise:
\begin{equation}
    \hat{P}_o(s, d) = \phi(s,d)^T \bm{\beta}, \quad P = \hat{P}_o(s, d) + \epsilon,
\end{equation}
where $\phi(\cdot)$ is the transformation to second degree polynomial space, $\bm{\beta}$ is the weight vector, and $\epsilon \sim \mathcal{N}(0, \sigma^2)$ is the observation noise, assumed to follow an independent, identically distributed Gaussian.

\begin{figure*} 
    \centering
    \subfloat[Throughput, $H_A: \mu < \mu_0$\label{throughput_less}]{%
       \includegraphics[width=0.33\textwidth]{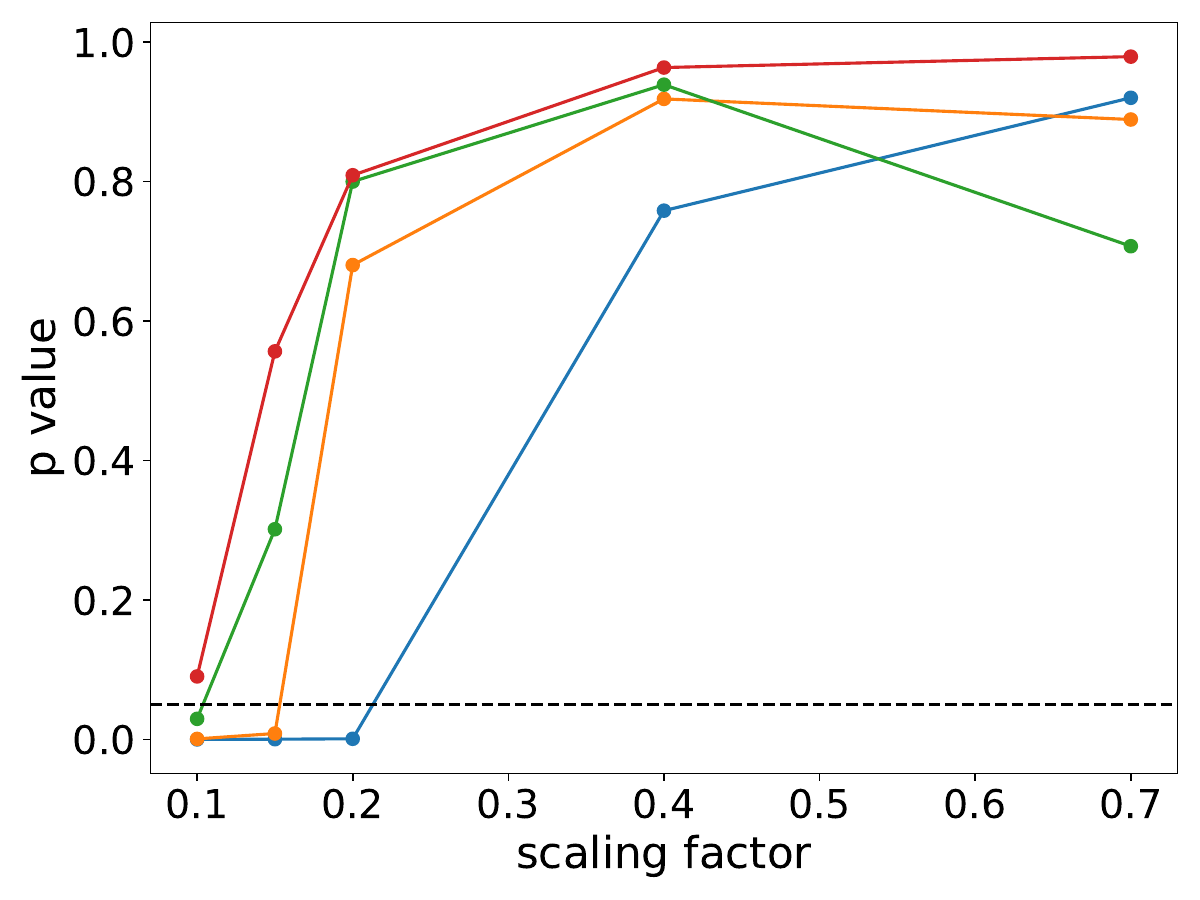}}
       \hfill
    \subfloat[Total Error, $H_A: \mu < \mu_0$\label{total_error_less}]{%
       \includegraphics[width=0.33\textwidth]{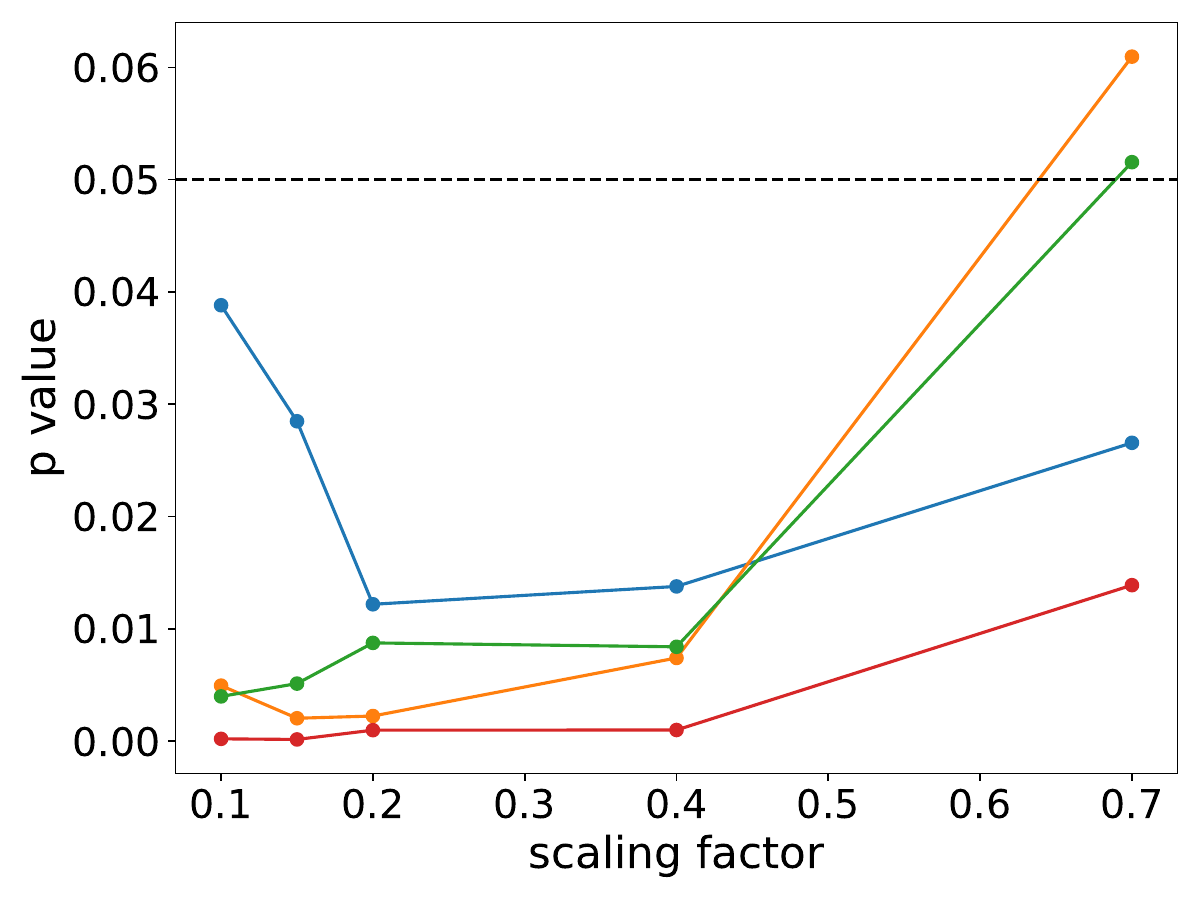}}
       \subfloat[Weighted Performance ($w$=0.5), $H_A: \mu > \mu_0$\label{weighted_performance_greater}]{%
       \includegraphics[width=0.33\textwidth]{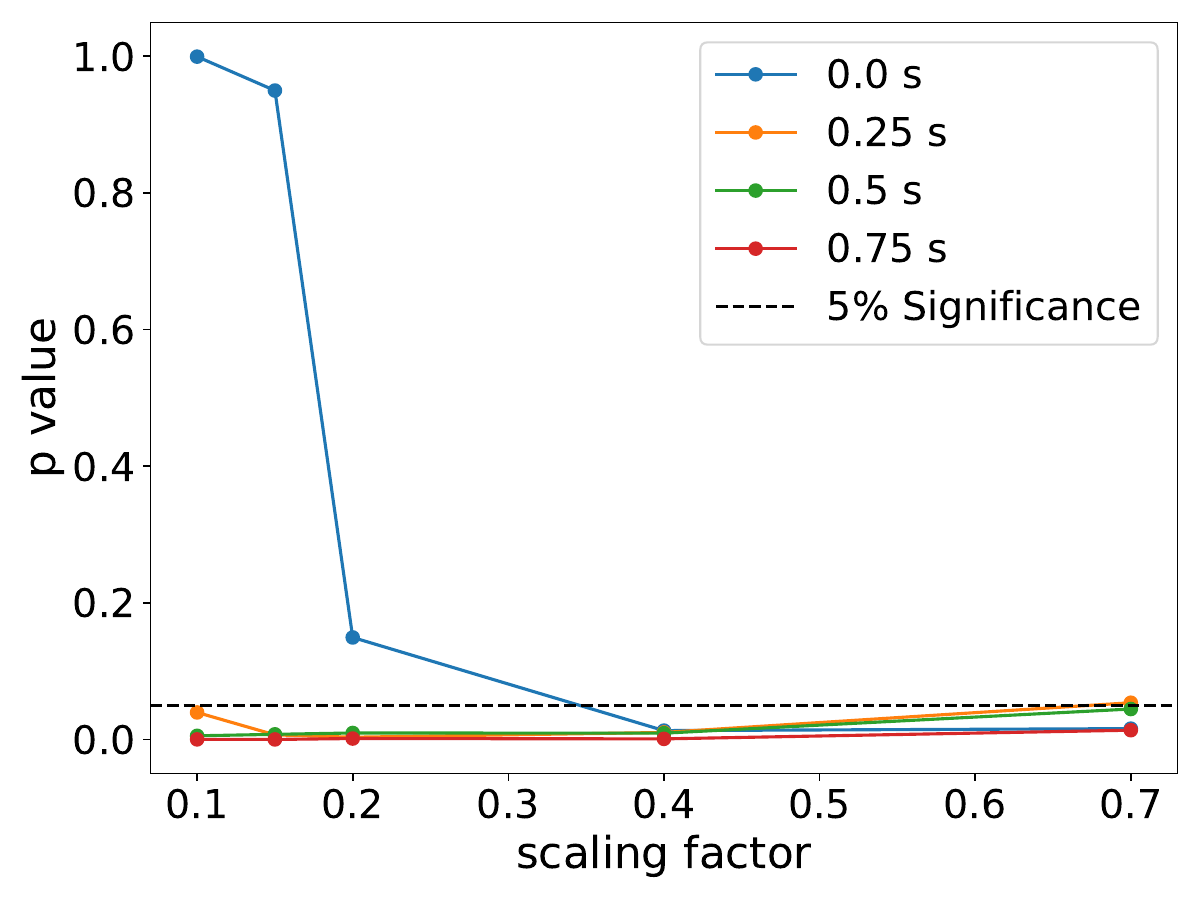}}
  \caption{Paired-sample t tests are conducted for each scaling factor against the nominal scaling factor of 1.0, demonstrating that reduced scaling factors lead to 1) a decrease in throughput only for the lowest scaling factors and lower delay, 2) a decrease in total error for lower scaling factors under delay, and 3) an increase in weighted performance for lower scaling factors when under delay.}
  \label{fig:stat_test} 
\end{figure*}

For Bayesian Analysis, we define a prior distribution that represents a prior belief for the model parameters, in this case $\bm{\beta}$ and $\sigma^2$.
We select the Normal-Inverse-Gamma distribution, as is typical for BPR, yielding
% It is common practice to choose a \textit{conjugate prior}, meaning a particular family of functions that leads to a posterior distribution of the same family with updated parameters. 
% A conjugate prior for the joint distribution of $\bm{\beta}$ and $\sigma^2$ is the \textit{normal-inverse-gamma} distribution. 
% If the distribution of $\bm{\beta}$ conditioned on $\sigma^2$ is a normal distribution with mean and variance $m$ and $V$
\begin{equation}
    \bm{\beta} | \sigma^2 \sim \mathcal{N} (\textbf{m}, \textbf{V})
\end{equation}
for the conditional distribution with $\textbf{m}$ and $\textbf{V}$ for the mean and variance, respectively, and
% the marginal distribution of $\sigma^2$ is the inverse-gamma distribution (a gamma distribution defined over the precision $\frac{1}{\sigma^2}$)
\begin{equation}
    \frac{1}{\sigma^2} \sim \Gamma(b/2, a/2)
\end{equation}
for the marginal distribution with $b/2$ and $a/2$ for the shape and rate, respectively, to define the Inverse-Gamma distribution.
The product of these distributions yields the joint distribution, which is also of the form Normal-Inverse-Gamma distribution with parameters $(\textbf{m}^*, \textbf{V}^*, b^*, a^*)$.
Given a dataset of $N$ samples, the joint distribution of the parameters are estimated as
\begin{equation}
    \textbf{m}^* = (\textbf{V}^{-1} + \textbf{X}^T \textbf{X})^{-1} (\textbf{V}^{-1} \textbf{m} + \textbf{X}^T \textbf{y})
\end{equation}
\begin{equation}
    \textbf{V}^* = (\textbf{V}^{-1} + \textbf{X}^T \textbf{X})^{-1}
\end{equation}
\begin{equation}
    a^* = a + \textbf{m}^T \textbf{V}^{-1} \textbf{m} + \textbf{y}^T \textbf{y} - (\textbf{m}^*)^T (\textbf{V}^*)^T \textbf{m}^*
\end{equation}
\begin{equation}
    b^* = b + N
\end{equation}
where $\textbf{X} = [s_{1:N}^\top, d_{1:N}^\top]$ and $\textbf{y} = P^\top_{1:N}$ are the dataset.
For inference and predicting performance, $P_o$, on a test point, $(s_o, d_o)$, the probability of performance is computed by marginalizing out the parameters, $(\textbf{m}^*, \textbf{V}^*, b^*, a^*)$.
The result is a Student's t-distribution of the form
\begin{equation}
    p(P | \textbf{X}, \textbf{y}, s, d) = t(d^*, [s, d] \textbf{m}^*, a^*(\textbf{I} + [s, d]^\top\textbf{V}^* \textbf{X})
    \label{eq:student_t_dist}
\end{equation}
where $t(\cdot)$ is the probability distribution function of the Student's t-distribution.
For a detailed derivation of the BPR equations, please see \cite{O’Hagan_Forster_Kendall_2004}.

% Since it is a conjugate prior, the posterior distribution for $\bm{\beta}$ and $\sigma^2$ after seeing training data $\{X, y\}$ with $n$ examples is also Normal-Inverse-Gamma with updated hyperparameters:

% \begin{equation}
%     m^* = (V^{-1} + X^T X)^{-1} (V^{-1} m + X^T y)
% \end{equation}
% \begin{equation}
%     V^* = (V^{-1} + X^T X)^{-1}
% \end{equation}
% \begin{equation}
%     a^* = a + m^T V^{-1} m + y^T y - (m^*)^T (V^*)^T m^*
% \end{equation}
% \begin{equation}
%     d^* = d + n
% \end{equation}

\subsection{Prior Distribution Modeling}

We consider two different approaches in selecting prior hyper-parameters, $(m, \textbf{V}, b, a)$, depending on if 1) there is no prior information or 2) there is prior data from other operators.

\subsubsection{Noninformative Prior}

When no reasonable assumptions can be made about the model parameters a priori, the standard practice is to use a non-informative prior. This corresponds to setting a zero mean, $\textbf{m} = 0$, and letting the prior variances tend to infinity by setting $\textbf{V}^{-1} = 0$ and $a = 0$.
To ensure a strictly positive scale parameter for $\sigma^2$, the shape is set to the number of feature dimensions, $b=-6$.
%It is often recommended for a strictly positive scale parameter such as $\sigma^2$ to use the improper prior distribution $f(\sigma^2) \propto \sigma^{-2}$, which corresponds to setting $b = -p$, where $p$ is number of feature dimensions, in our case 6.
%It can be shown that using such a prior leads to a posterior $m^*$ equal to the coefficients obtained from the Maximum Likelihood Estimate (MLE), $\hat{\bm{\beta}}$. Thus, the predictions made under this approach are the same as that of the MLE approach.

\subsubsection{Informative Prior}
In the case where prior information is available through other users' data, we use that data to inform the prior distribution.
We accomplish this by solving for the hyper-parameter set which maximizes the other users' data probability, i.e. Maximum Likelihood Estimation.
Formally, we solve for the prior hyper-parameters
\begin{equation}
    \argmax_{\textbf{m}, \textbf{V}, \textbf{a}, \textbf{b}} \prod_{i=1}^M p( P_i, \textbf{X}_i, \textbf{y}_i)
\end{equation}
where $P_i, \textbf{X}_i, \textbf{y}_i$ for $i=1,\dots,M$ are the other users' datasets.
% This can be done by first calculating the estimated model parameters $\hat{\bm{\beta}}_i$ and $\hat{\sigma}_i^2$ for each individual in the prior datasets using MLE. Then, treating $\hat{\bm{\beta}}_i$ and $\hat{\sigma}_i^2$ as samples from the normal and inverse-gamma distributions, respectively, the hyperparameters $(\hat{m}, \hat{V}, \hat{d}, \hat{a}$ can be inferred, again using MLE, to use for the prior. 

\subsection{Optimizing for the Scaling Factor}

Once the operator model in $\eqref{eq:student_t_dist}$ is solved for, we find the optimal scaling factor $s^*$ depending on the delay for the operator.
Formally, we express the optimization problem as
\begin{equation}
    s^*(d) \, = \, \argmin_s \; \hat{P}_o(s, d),
\end{equation}
which is solved by directly searching for the maximum value across a discrete set of scale values.

\section{Simulation Experiments}

A group of $n = 10$ participants were selected for the first user study to measure the performance of teleoperative tasks under different levels of latency and scaling factors in a simulated environment. The participants were all non-experts, with no prior experience in telesurgery. 
A 2D target acquisition task was modeled after the Fitt's Law paradigm (commonly used to evaluate human-computer interfaces and teleoperative devices), in which the computer mouse acts as the leader, with a rendered "instrument" on screen as the follower, as shown in Fig. \ref{fig:cover_fig}.

\subsection{Procedure}

Participants first completed a five-minute practice session to become familiar with the controls, which included using the spacebar to toggle the clutch, which decouples the motion between leader and follower, mimicking the controls of common telesurgical systems. For each trial, users started by clicking the center of the first target (at which point the timer starts) and continued clicking each subsequent target as quickly and accurately as possible. The users were instructed that they are scored on both time and accuracy, and that each click triggers the next target (users were told not to attempt to click a target more than once). Users performed a total of 24 trials for every combination of scaling factor in the set $(0.1, 0.15, 0.2, 04, 0.7, 1.0)$ and delay in $(0, 0.25, 0.5, 0.75)$ seconds. The range of delay values were chosen to span a range from no delay, up to the point where previous research has suggested performance should be significantly affected \cite{anvari2007remote, wang2024influence, anvari2005impact, marescaux2002transcontinental, xu2014determination, barba2022remote}. The set of scaling factor values were chosen so that users were forced to use the clutch mechanism at the lower scaling factors and could easily span the target space at the higher scaling factors. The order of scaling factor and delay combinations were randomized for each user.

\begin{figure*}[t]
    \centering
    \includegraphics[width=0.99\textwidth]{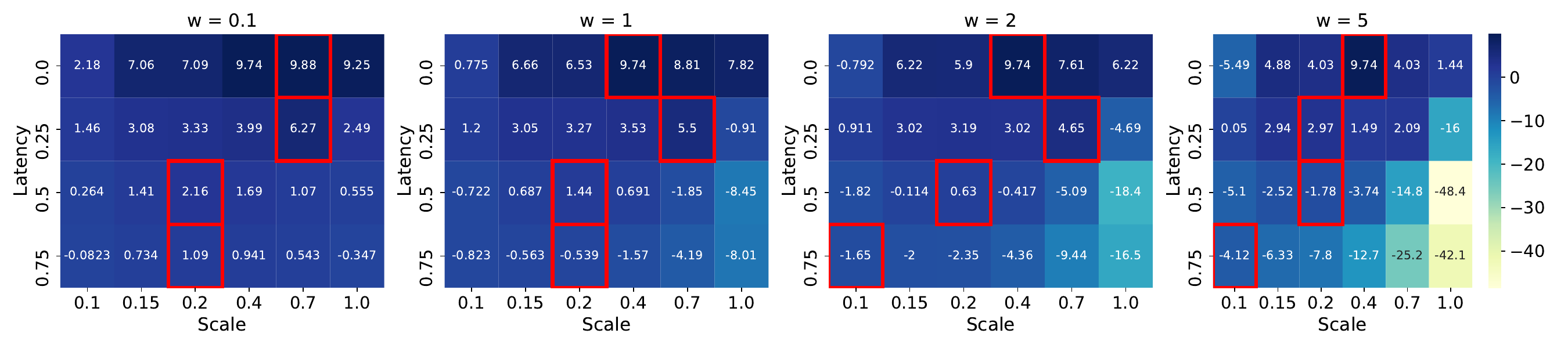}
    \caption{Heatmaps of the weighted performance metric for one user with different values of $w$ demonstrate how changing the relative weight of throughput and error leads to different optimal scaling factors. As $w$, the priority for safety, is increased, the optimal scaling factor tends to decrease for all latency levels.
    }
    \label{fig:WP}
\end{figure*}

\subsection{Evaluating Performance}

In surgical scenarios, speed and safety are important factors to consider but are inversely related; surgeons want to perform as quickly as possible, without being unsafe.
Task completion time measures speed, but cannot be compared across tasks of varying difficulty. Therefore we use \textit{throughput}, a common metric used to evaluate performance of human-computer interfaces and teleoperative devices \cite{soukoreff2004towards}. Throughput depends on both completion time and task difficulty. It is mathematically defined as

\begin{equation}
    TP = \frac{log_2 (\frac{D}{W} + 1)}{T},
\end{equation}
where $T$ represents the average time to move between targets. The numerator represents the "index of difficulty", with $D$ defined as the distance between targets and $W$ as the width of the target.
By essentially normalizing the completion time by the degree of task difficulty, throughput provides a way to compare the speed aspect of performance across tasks of varying difficulties.

In order to measure safety, we rely on several error metrics: $\Delta D$, the average deviation from the target center,  and $OSD$, a metric measuring the amount of overshoot for each target. Overshoot is one of the main issues encountered in teleoperation under delay, and in order to quantify it we define $OSD$ as
\begin{equation}
    OSD = \int max(\dot{r}, 0) dt,
\end{equation}
where $\dot{r}$ is the velocity of the instrument relative to the intended target. $OSD$ can be interpreted as the cumulative distance for which the instrument moves \textit{away} from the target center.

Finally, since surgical performance must be a balance between speed and safety, we introduce a "weighted performance" metric that is the weighted sum of speed and safety metrics:

\begin{equation}
    WP = (1-w) * (TP) - w * (OSD + \Delta D),
\end{equation}
where $w$ is a weight that can be adusted to vary the priority of speed or safety.

\subsection{Results}

% \begin{figure} 
%     \centering
%   \subfloat[User A\label{user_a}]{%
%        \includegraphics[width=0.99\linewidth]{Figures/heatmap_key_metrics_sujaan.png}}
%     \\
%   \subfloat[User B\label{user_b}]{%
%         \includegraphics[width=0.99\linewidth]{Figures/heatmap_key_metrics.png}}
%   \caption{Values of key performance metrics vs. latency and scale for two users in the physical study.}
%   \label{fig:key_metrics_physical}
% \end{figure}

Results from both simulated and physical experiments provide evidence confirming the following hypotheses:
\begin{enumerate}
    \item The relationship between performance, latency, and delay is user-specific. Each users' personal tendencies influence what scaling factor they will perform best with at different levels of latency.
    \item In general, a lower scaling factor leads to decreased error when latency is present.
    \item In terms of overall performance, most users prefer lower scaling factors as latency increases.
\end{enumerate}

Hypothesis 1) is apparent in Fig. \ref{fig:key_metrics_sim}, which shows a heatmap of several key performance metrics as a function of scaling factory and latency.
The optimal scaling factor as measured by these metrics changes across latency levels and between the two users, which motivates the need for personalized models in order to achieve the best performance for each user.

In Fig. \ref{fig:optimal_scale_vs_latency}, we observe that the optimal scaling factor as measured by the weighted performance metric varies among users at a given latency and generally decreases as latency increases. This trend suggests that adaptive scaling strategies, which dynamically adjust based on real-time latency, could be key to optimizing performance in telesurgical applications.

A set of paired-sample t tests are conducted to show hypotheses 2) and 3). Changing the scaling factor leads to statistically significant changes in the performance metrics, as shown in Fig. \ref{fig:stat_test}. Fig \ref{throughput_less} shows that throughput has a statistically significant decrease at the lowest scaling factors, for all except the highest delay of 750 ms. This aligns with previous research that lower scaling factors can lead to decreased task completion speed \cite{richter2019motion}, but this may not be true for high delays in which overshoot and oscillations are more significant. Fig. \ref{total_error_less} shows that the total error is reduced for scaling factors of 0.4 and below, for all levels of delay (with the exception of zero delay at the lowest scaling factors). This confirms that lowering the scaling factor leads to a decrease in error under delay. These results indicate there is a trade-off between the increase in safety and reduction in speed with lower scaling factors.

This is why it helps to define a weighted performance metric that combines speed and safety. As Fig. \ref{weighted_performance_greater} shows, the weighted performance metric (when throughput and error are evenly weighted) has a statistically significant increase for scaling factors below 0.5 at all levels of delay, with the exception of zero delay. This highlights the power of motion scaling; although it may decrease speed slightly, the advantage gained by reducing the amount of overshoot and oscillations under delay can lead to an increase in overall performance.
One can also vary the ratio $w$ to adjust the priority for safety or speed, as depicted in Fig. \ref{fig:WP}. For lower values of $w$ (higher priority for speed), the optimal scaling factors are higher, and as the priority for safety is increased, the optimal scaling factor decreases for all levels of delay.

% \begin{figure}
%     \centering
%     \subfloat[Original Data]{\includegraphics[width=0.49\linewidth]{Figures/original_data_heatmap.pdf}}
%     \subfloat[Prediction Over Dense Input]{\includegraphics[width=0.49\linewidth]{Figures/dense_pred_heatmap.pdf}}
%     \caption{Original data and model predictions over a dense input demonstrate the model's ability to interpolate the relationship between performance metrics and latency/scale for untested input parameters. Green circles denote points used for training, and the optimal scaling factor per latency is highlighted in red.}
%     \label{fig:model1_interpolation}
% \end{figure}

\subsection{Modeling Results}

\begin{figure}
    \centering
    \includegraphics[width=0.99\linewidth]{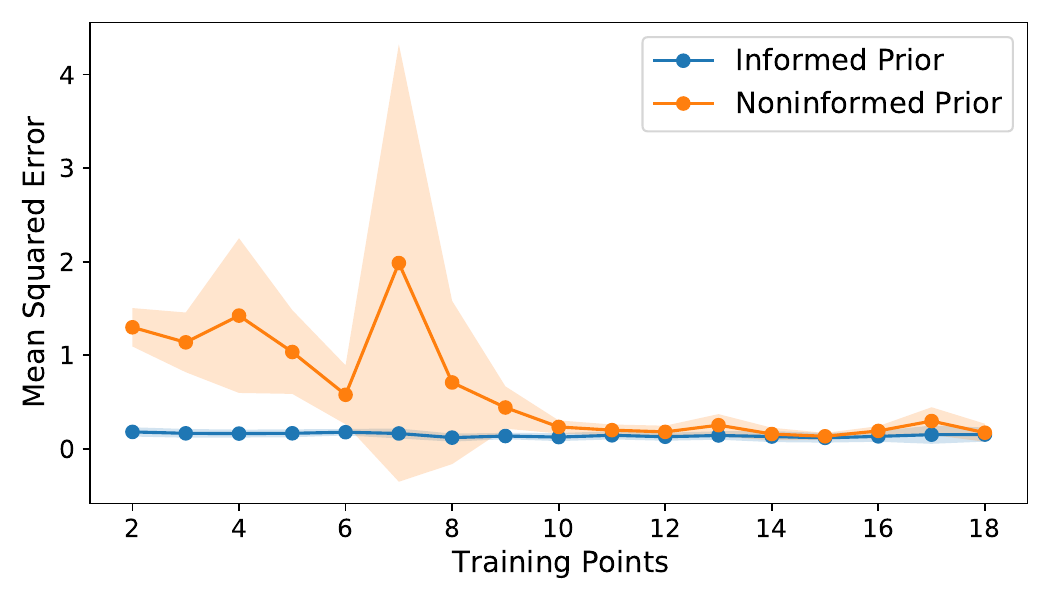}
    \caption{Comparing the mean squared error for informed vs. noninformed prior illustrates the power of leveraging prior data in a Bayesian Regression Model. The informed prior for a given user is estimated from the datasets of the other nine users, leading to better accuracy with fewer training points.}
    \label{fig:comparing_priors}
\end{figure}

In order to evaluate the model's performance, the mean squared error is computed for a range of training set sizes.
Fig. \ref{fig:comparing_priors} shows that incorporating other users’ data as a prior in the regression model significantly improves prediction accuracy with fewer training points. This finding underscores the potential of leveraging collective user data to enhance individual performance predictions, thus paving the way for more efficient and personalized telesurgical systems.

% \subsection{Demographic Analysis}

% \begin{itemize}
%     \item We also perform statistical significance tests to asses whether any differences in performance are present between different demographic groups.
%     \item \textit{TABLE FOR DEMOGRAPHIC RESULTS}
%     \item This analysis shows...
% \end{itemize}

% \begin{figure}
%     \centering
%     \includegraphics[width=0.9\linewidth]{Figures/demographics.PNG}
%     \caption{Different demographic groups perform differently or the same...}
%     \label{fig:demographics}
% \end{figure}

\section{Physical Experiments}

A user study is also conducted using a physical surgical robot to demonstrate the methods in a more realistic surgical setting. 
The experiments are conducted using the da Vinci Research Kit \cite{kazanzides2014open} in teleoperation mode. The MTM (Master Tool Manipulator) console is connected locally to the PSM (Patient Side Manipulator) arms and ECM (Endoscopic Camera Manipulator), and the delay is simulated using a buffer between MTM arms and PSM console. The stereo camera information is not delayed, and thus the delay set by the buffer can be considered the round trip delay under the assumption delay is equal in both directions. 

\subsection{Procedure}
The task for these experiments, depicted in FIg. \ref{fig:cover_fig} is inspired by the standard Peg Transfer task from the Fundamentals of Laparoscopic Surgery Manual Skills training program \cite{fls}, in which the user picks up each object (in any order), passes it to the other hand, places it on one of the marked pegs on the other side, and then repeats the process in reverse.
The participants complete 8 trials with latency at 0.2 seconds and 0.5 seconds, and scaling factor at 0.1, 0.2, 0.3, 0.4. These latency values are chosen because 0.2 seconds has been reported as the approximate level at which surgeons start to feel impeded by the delay, and 0.5 seconds should be well past the limit where surgeons feel impeded. Thus the experiments are conducted at one latency levels where the effects of delay are influential but not severe, and one level where the effects should be significantly influential. The scaling factors are chosen as a spread around the nominal defualt scaling factor for the dvrk, 0.2. 

The participants are given basic instructions on the task and operating controls. They then complete four full training trials at zero latency and nominal scaling (0.2), zero latency and 0.1 scaling, 0.5 seconds latency and 0.1 scaling, and 0.5 seconds latency and 0.4 scaling. The training is intended to remove the learning curve factor from the data. 

\subsection{Evaluating Performance}

The metrics used to evaluate performance for these experiments are again meant to capture both speed and safety. We can use the metric of throughput again, where $W$ is now defined as the width of the hole in the object minus the width of the peg, and D is the average distance between pegs. Since it is possible for a user to drop an object outside the field of view of the camera, $T$, the average time to move between targets, is computed as the total completion time divided by the number of successful transfers.

The FLS standard is to only count the number of dropped objects as an error metric; however, this alone does not capture a lot of the information that affects performance. For example, as discussed before, overshoot is one of the biggest issues introduced by delay, and this may not necessarily correlate to a higher number of dropped objects. As a way to estimate the amount of overshoot, an ATI Axia80-M20 force sensor is placed underneath the peg board to measure the force imparted to the peg board. The wrench force of the PSM arms is also measured, and the time-averaged sum of these forces over the trial is used as an error metric.

% \begin{figure}
%     \centering
%     \includegraphics[width=\linewidth]{Figures/peg_board.pdf}
%     \caption{In the physical experiments, users complete a peg transfer task inspired by the official FLS training program. The average time per transfer, force imparted to the pegboard, and forces experienced by the PSM arms are used as performance metrics.}
%     \label{fig:pegtransfer}
% \end{figure} 

\subsection{Results}

\begin{figure}
    \centering
    \subfloat[Throughput]{
    \includegraphics[width=0.99\linewidth]{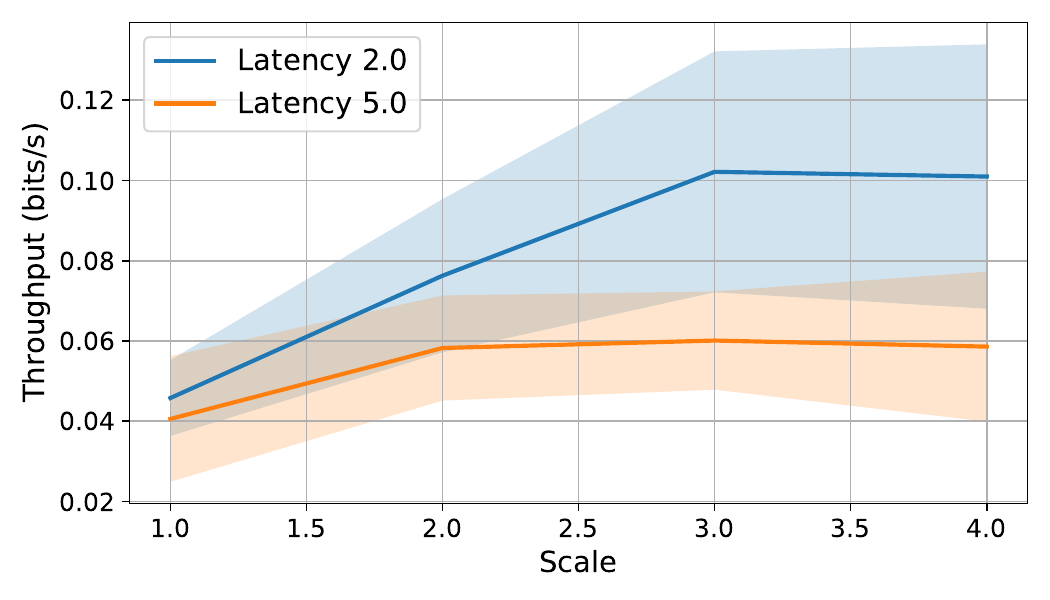}
    }
    \\
    \subfloat[Force Penalty]{
    \includegraphics[width=0.99\linewidth]{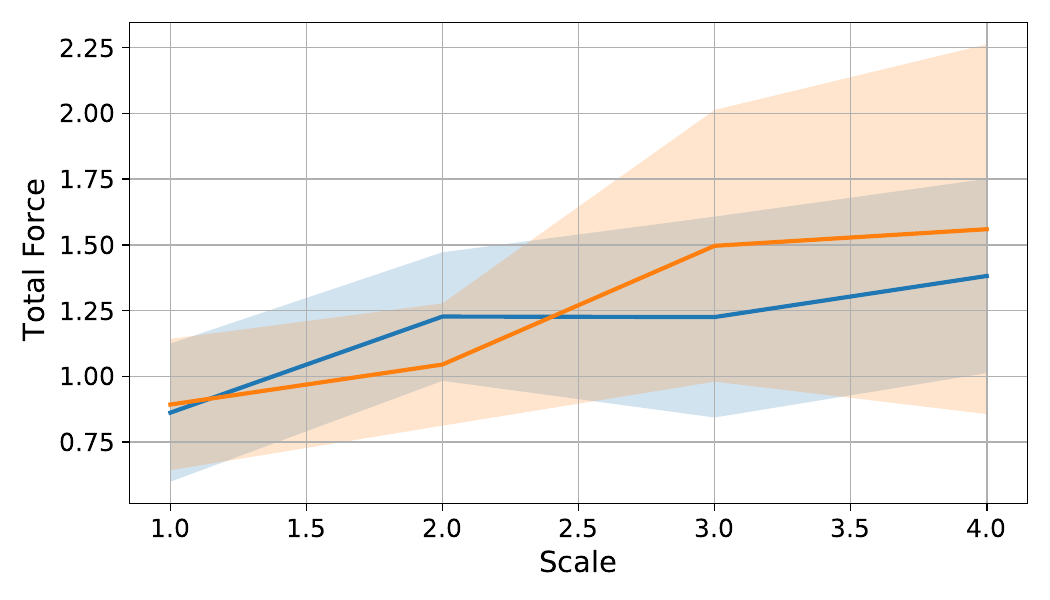}}
    \caption{Mean and standard deviation plots of the throughput and force metrics demonstrate the same trends as found by the 2D user study: speed tends to increase as scaling factor increases, but at the cost of safety (higher force metric). This highlights the tradeoff between speed and safety.}
    \label{fig:dvrk_results}
\end{figure}

The results of the physical experiments largely mirrored those observed in the 2D simulation experiments. Specifically, lower scaling factors led to lower throughput, and error increased with increasing scaling factor, as shown in Fig \ref{fig:dvrk_results}. This consistency suggests that the trends observed in the simulated environment translate well to real-world scenarios.

The experiments also revealed that higher latency values generally resulted in longer completion times. However, unlike the 2D simulations, the impact of latency on the error rate was less pronounced in the physical experiments. This discrepancy could potentially be attributed to differences in the complexity of the tasks.

\begin{table}[h!]
    \centering
    \caption{Two-Way ANOVA Results for Physical Experiments}
    \begin{tabular}{@{}lcccc@{}}
        \toprule
        \textbf{Effect} & \textbf{Sum of Squares} & \textbf{df} & \textbf{F Value} & \textbf{p-value} \\ 
        \midrule
        \multicolumn{5}{c}{\textit{Throughput}} \\
        C(latency)         & 0.0184   & 1   & 25.46  & 3.00e-06 \\
        C(scale)           & 0.0219   & 3   & 10.08  & 1.30e-05 \\
        C(latency):C(scale) & 0.0081  & 3   & 3.73   & 0.0149   \\
        Residual           & 0.0513   & 71  &        &          \\ 
        \midrule
        \multicolumn{5}{c}{\textit{Force Penalty}} \\
        C(latency)         & 0.11    & 1   & 0.68   & 0.4119   \\
        C(scale)           & 4.04    & 3   & 8.29   & 0.0001   \\
        C(latency):C(scale) & 0.59    & 3   & 1.20   & 0.3154   \\
        Residual           & 11.53   & 71  &        &          \\ 
        \bottomrule
    \end{tabular}
    \label{tab:anova_results}
\end{table}

\subsubsection{Statistical Tests}

A two-way Analysis of Variance (ANOVA) was performed to evaluate the effects of latency and scale, as well as their interaction, on the performance metrics \textit{throughput} and \textit{force penalty}. The results indicate that both latency and scale significantly influence these metrics, while the interaction effect was not statistically significant.

The ANOVA for \textit{throughput} revealed significant main effects for both latency ($F(1, 71) = 25.46, p < 0.001$) and scale ($F(3, 71) = 10.08, p < 0.001$). Higher latency was associated with lower throughput, indicating that increased delay negatively impacts task efficiency. Similarly, changes in scale significantly influenced throughput, suggesting that task performance is sensitive to the level of precision required. The interaction effect between latency and scale was significant ($F(3, 71) = 3.73, p = 0.015$), indicating that the effect of scale on throughput depends on the level of latency. This was supported by variations in the trend of throughput across different latency conditions, suggesting that an optimal scale setting may depend on the specific latency environment.

For \textit{force penalty}, scale had a significant main effect ($F(3, 71) = 8.29, p < 0.001$), with larger scales (lower precision) resulting in greater penalties. In contrast, latency did not significantly influence \textit{force penalty} ($F(1, 71) = 0.68, p = 0.41$), and the interaction term was also not significant ($F(3, 71) = 1.20, p = 0.32$). These findings suggest that in these experiments, the force penalty and amount of overshoot are primarily governed by scale, with little to no modulation by latency.

% The ANOVA model for \textit{throughput} explained 51.89\% of the variance ($R^2 = 0.5189$), indicating a moderate fit, while the model for \textit{force penalty} explained only 29.10\% ($R^2 = 0.2910$). This highlights that while latency and scale are key drivers of task completion time, other unmodeled factors may significantly impact force control.

\section{Discussion and Conclusion}

This study confirms previous research that motion scaling is a simple and effective strategy to mitigate latency effects in telesurgical tasks. The results from both simulated and physical environments illustrate that the influence of motion scaling on performance varies not only with the level of latency but also among individual users. These findings emphasize the importance of developing personalized models to optimize the advantages offered by motion scaling for teleoperative systems.

Our experiments highlight a fundamental trade-off between speed and safety. Higher scaling factors enable faster task completion but increase error rates, particularly under high-latency conditions. Conversely, lower scaling factors enhance precision and reduce errors at the expense of slower task execution. These results suggest that the optimal scaling factor depends on the performance priorities—speed or safety—of a given task. Combining these metrics into a weighted sum allows for the identification of a scaling factor that optimally balances speed and safety. Notably, this optimal factor tends to decrease with increasing latency, indicating that lower scaling is generally more beneficial for overall performance in high-latency scenarios.

Furthermore, the relationship between performance, scaling factor, and latency is inherently user-specific. We demonstrate that simple regression models can effectively capture this relationship for individual users, enabling predictions of optimal scaling factors across various latency levels. Using Bayesian Regression further we can leverage data from other users as a prior, significantly improving prediction accuracy with minimal training data. This approach holds promise for real-world systems, for which a set of global prior hyperparameters can be developed and applied to new users, avoiding the need to start from scratch.

Physical experiments conducted with the da Vinci Research Kit \cite{kazanzides2014open} validate the findings from simulated environments, showing that trends observed in 2D tasks also translate to realistic 3D telesurgical scenarios. These results underscore the robustness of our approach and its applicability to complex surgical tasks.

Future research should focus on refining these predictive models and evaluating their applicability in more complex surgical tasks. The extension of these methods to other teleoperation domains, such as space exploration and remote manufacturing, also holds significant potential. By advancing the feasibility of telesurgery and other teleoperative technologies, this research contributes to bridging critical gaps in healthcare access, particularly in remote and underserved areas. Ultimately, these developments aim to make advanced medical procedures safer, more efficient, and universally accessible.

\section{Acknowledgements}
This work was supported in part by the US Army Medical Research and Development Command and the US Army SBIR Program under contract W81XWH19C0096.

% \clearpage
\balance
\bibliographystyle{ieeetr}
\bibliography{refs}
\balance

\end{document}